\documentclass{article}

\usepackage[nonatbib,preprint]{neurips_2021}
\usepackage{wrapfig}
\usepackage[utf8]{inputenc} % allow utf-8 input
\usepackage[T1]{fontenc}    % use 8-bit T1 fonts
\usepackage[hidelinks]{hyperref}       % hyperlinks
\usepackage{url}            % simple URL typesetting
\usepackage{booktabs}       % professional-quality tables
\usepackage{amsfonts}       % blackboard math symbols
\usepackage{nicefrac}       % compact symbols for 1/2, etc.
\usepackage{microtype}      % microtypography
\usepackage{xcolor}         % colors

\usepackage{amsmath}
\usepackage{soul}

\title{An Explainable Probabilistic Classifier for Categorical Data Inspired to Quantum Physics}

\author{%
  Emanuele Guidotti \\
  Institute of Financial Analysis\\
  Université de Neuchâtel, Switzerland\\
  \texttt{emanuele.guidotti@unine.ch} \\
  \And
  Alfio Ferrara \\
  Department of Computer Science and Data Science Research Center\\
  Università degli Studi di Milano, Italy \\
  \texttt{alfio.ferrara@unimi.it} \\
}

\begin{document}

\maketitle

\begin{abstract}
This paper presents Sparse Tensor Classifier (STC), a supervised classification algorithm for categorical data inspired by the notion of superposition of states in quantum physics. By regarding an observation as a superposition of features, we introduce the concept of wave-particle duality in machine learning and propose a generalized framework that unifies the classical and the quantum probability. We show that STC possesses a wide range of desirable properties not available in most other machine learning methods but it is at the same time exceptionally easy to comprehend and use. Empirical evaluation of STC on structured data and text classification demonstrates that our methodology achieves state-of-the-art performances compared to both standard classifiers and deep learning, at the additional benefit of requiring minimal data pre-processing and hyper-parameter tuning. Moreover, STC provides a native explanation of its predictions both for single instances and for each target label globally. All the code is released at \texttt{\href{https://sparsetensorclassifier.org}{https://sparsetensorclassifier.org}}
\end{abstract}

\section{Introduction}\label{sec:intro}
The application of supervised learning algorithms for data classification in an increasing number of different contexts requires novel machine learning methods capable of improving generalisability and delivering performances robust to data pre-processing, parameter and hyper-parameter tuning, and small samples~\cite{kelly2019key, lo2018machine, vabalas2019machine}. In particular, feature extraction and data pre-processing are crucial steps for categorical data, which are often highly heterogeneous in terms of size, structural differences, and noise. Representing this kind of data in the feature space requires a task-specific feature engineering work that is non-trivial and time consuming.
Another limitation, that can potentially hinder the adoption of machine learning methodologies, is the growing demand for explainable and interpretable classification models, both for what concerns the classification of single instances (local explanation) and the rationale of a model prediction in terms of the most relevant features for a target class label (global explanation). 
In this context, we present Sparse Tensor Classifier (STC), an explainable classification algorithm for categorical data inspired by the notion of superposition of states in quantum physics. The empirical evaluation against consolidated algorithms and deep learning alternatives demonstrates the native capability of STC to achieve state-of-the-art performances without relying on data pre-processing and (hyper-)parameter tuning, at the additional benefit of providing a meaningful explanation of the classification results.

\paragraph{Related work.}\label{sec:relwork} 
Several  machine  learning  algorithms have long been studied in the context of categorical data in general and text classification in particular (see \cite{korde2012text} for a survey). 
This extensive literature includes simple but effective methods for text categorization with multi-label documents and many categories (see e.g., \cite{popa2007text}) and, more recently, deep learning alternatives based on Knowledge Base descriptors \cite{yamada2018}, cooperative Neural Networks \cite{shrivastava2018}, Neural Attentive Bag-of-Entities Models \cite{yamada2019}, ensemble Neural Networks \cite{zhang2020}. 
In all these approaches, feature engineering and data pre-processing are crucial steps given the unstructured and noisy nature of text \cite{kowsari2019text}. The need of dealing with many and potentially noisy categorical features goes beyond textual applications and it is shared across different disciplines. As an example, Extended Connectivity Circular Fingerprints \cite{rogers2010extended} are used to compute a bag-of-words style representation of a molecule for machine learning applications in cheminformatics and drug discovery \cite{lo2018machine}.
Another field related to our work is machine learning explanation and interpretation. Although there is still no agreed-upon formal definition of \textit{interpretability} (see \cite{lipton2018mythos,DBLP:conf/icml/ChenSWJ18}), a taxonomy of explanation methods is given in~\cite{arya2019one}. With respect to this framework, STC provides a native explanation of its result that is both `local' and `global', in that we compute a score to determine what features are important for a certain class both for single predictions (i.e., local explanation) and for the entire model (i.e., global explanation). 
In this paper, we also rely on LIME~\cite{ribeiro2016should} to compute explanations from a variety of other algorithms in order to compare them with STC. LIME learns an interpretable model, such as a linear regression model, in the proximity of each prediction and computes importance values based on the coefficients in the interpretable model. We refer the reader to the survey in~\cite{DBLP:conf/icml/AnconaOG19,arya2019one,pmlr-v119-moshkovitz20a} for more explanation methods. 

\paragraph{Contribution.}
By regarding an observation as a superposition of features, we introduce the concept of wave-particle duality in machine learning and propose a generalized framework that unifies the classical and the quantum probability. Embracing the new heuristic can be crucially beneficial for machine learning applications to better describe reality, similarly to what Einstein stated about the dual nature of light back in 1938 \cite{einstein1938}:

\begin{quote}
It seems as though we must use sometimes the one theory and sometimes the other, while at times we may use either. We are faced with a new kind of difficulty. We have two contradictory pictures of reality; separately neither of them fully explains the phenomena of light, but together they do.
\end{quote}

On the practical side, we develop a new supervised classification algorithm that implements the wave-particle duality for categorical data, together with an entropy-based mechanism for noise filtering. We show that our methodology achieves state-of-the-art performances compared to both standard classifiers and deep learning, at the additional benefit of requiring minimal data pre-processing and hyper-parameter tuning. Moreover, the proposed algorithm supports multiclass and multilabel classification, online learning, prior knowledge, automatic dataset balancing, and provides a native explanation of its predictions both for single instances and for each target class label globally. 

The paper is organized as follows: Section~\ref{sec:pre} presents some preliminary notions required to develop our methodology described in Section~\ref{sec:theory}. Section \ref{sec:algorithm} presents the STC algorithm whose standard configuration is empirically evaluated in Section~\ref{sec:test}. Finally, Section~\ref{sec:conc} gives our concluding remarks.
%The paper is organized as follows: in Section~\ref{sec:pre}, we present some preliminary notions concerning the STC approach. In Sections~\ref{sec:theory} and \ref{sec:algorithm}, we discuss how STC performs learning and prediction from data. In Section~\ref{sec:test}, we provide out empirical evaluation of STC and, finally, in Section~\ref{sec:conc}, we give our concluding remarks.

\section{Preliminaries}\label{sec:pre}
We use the multi-index notation $i=(i_0,i_1,\dots, i_n)$ where the index $i$ is an ordered tuple of indices $(i_0,i_1,\dots, i_n)$. In this notation, $T_{ij}$ does not represent a matrix but rather a tensor of arbitrary dimensions $T_{i_0\dots i_nj_0\dots j_m}$. We use the index $i$ to identify the dimensions associated with the \emph{target} variable(s), and the index $j$ for the \emph{features}. We represent an \emph{observation} with the joint probability distribution $X_{ij}\equiv P_X(i,j)$ of the features and the target variables. The probability distribution can be arbitrarily chosen if prior knowledge is available or computed from the data by counting the number of co-occurrences. 
Then, given a set of observations $\{X^{(n)}_{ij}\}_{n=1,...,N}$ we compute the \emph{corpus} represented by the tensor:
\begin{equation}\label{eq:corpus}
	C_{ij} = \sum_{n} X^{(n)}_{ij}
\end{equation}

\paragraph{Classical probability.} 
The probability distribution of $i$ can be recovered from the distribution of $j$ by Bayes' theorem:
\begin{equation} \label{eq:bayes}
	X_i = P_X(i) = \sum_j P_X(i\mid j) P_X(j) 
\end{equation}

\paragraph{Quantum probability.}
In quantum physics a system is regarded as a superposition of states $s$ and, using Dirac's notation, is represented by the wave function $|\psi\rangle$:
\begin{equation} \label{eq:wave}
	|\psi\rangle = \sum_{s} \psi_{s} |s\rangle
\end{equation}
In the Copenhagen interpretation, the modulus squared of the inner product is interpreted as the probability of the wave function $\psi$ collapsing to a new wave function $\varphi$ upon measure of an observable. This is known as the Born rule, and is one of the fundamental postulates of quantum mechanics.
\begin{equation} \label{eq:born}
	|\langle\phi|\psi\rangle|^2 = \Bigl|\sum_s\phi^*_s\psi_s\Bigl|^2 = P(\psi\rightarrow \phi)
\end{equation}
From equations \eqref{eq:wave} and \eqref{eq:born}, we notice that\footnote{As long as we are using an orthonormal basis $\langle s|s'\rangle=\delta_{ss'}$} the coefficients $|\psi_{s}|^2$ represent the probability of observation $|\psi\rangle$ collapsing to state $|s\rangle$, that we denote with $X_{s}$.
\begin{equation}
	P(\psi \rightarrow s) = 
	|\langle s|\psi\rangle|^2 = \Bigl |\sum_{s'}\psi_{s'}\langle s|s'\rangle\Bigl |^2 = |\psi_{s}|^2 \equiv X_{s}
\end{equation}
Therefore the coefficients $\psi_{s}$ are given by the complex numbers:
\begin{equation}\label{eq:phase}
	\psi_{s} = e^{i\theta_s}\sqrt{X_{s}}
\end{equation}
where $\theta_s$ represents the phase of the coefficient $\psi_s$.

\section{Methodology}\label{sec:theory}
We now develop our methodology under the classical and the quantum probability settings, that are generalized and unified by the STC algorithm in Section \ref{sec:algorithm}.

\paragraph{Classical probability.}
In the classical probability setting, we estimate the conditional probability $P_X(i\mid j)$ in equation \eqref{eq:bayes} with the conditional probability $C_{i\mid j}$ computed from the corpus.
\begin{equation} \label{weight:classic}
	C_{i\mid j} = \frac{C_{ij}}{\sum_{i'} C_{i'j}}
\end{equation}
Given a new observation $X_j$ for which only the features $j$ are observed, we are now able to compute the probability distribution of the targets $i$ by Bayes' theorem:
\begin{equation} \label{eq:classic}
	X_i = \sum_j C_{i\mid j} X_j 
\end{equation}

\paragraph{Quantum probability.}
In the quantum probability setting, we regard an observation as a superposition of features:
\begin{equation} 
	|\psi\rangle = \sum_{j} \psi_{j} |j\rangle
\end{equation}
Consider a new observation $X_j$ for which only the features $j$ are observed. The probability distribution of the targets $i$ can be obtained by computing the probability of the wave function $|\psi\rangle$ to collapse on the targets $|i\rangle$:
\begin{equation}\label{eq:xi}
    X_i =
	|\langle i|\psi\rangle|^2 = 
	\Bigl |\sum_{j}\psi_{j}\langle i|j\rangle\Bigl |^2 =
	\Bigl|\sum_j \psi_j i^*_j\Bigl|^2
\end{equation}
where $\psi_j=e^{i\theta_j}\sqrt{X_{j}}$ as given in \eqref{eq:phase} and $i_j^*$ are the complex conjugates of the coefficients of the wave function $|i\rangle$ that represents the targets.
\begin{equation}
	|i\rangle = \sum_j i_j|j\rangle
\end{equation}
We obtain the coefficients $i_j$ by identifying the transition probability from $|i\rangle$ to $|j\rangle$ with the conditional probability of $j$ given $i$ computed from the corpus $C_{ij}$.
\begin{equation}\label{eq:ij}
    C_{j\mid i} = \frac{C_{ij}}{\sum_{j'} C_{ij'}} =
	P(i\rightarrow j) = |\langle j | i \rangle|^2 =
	|i_j|^2
\end{equation}
From equation \eqref{eq:ij} we get the representation:
\begin{equation}
i_j=e^{i\phi_{ij}}\sqrt{C_{j\mid i}}
\end{equation}
where $\phi_{ij}$ represents the phase of the coefficient $i_j$. We are now able to compute the probability distribution of the targets $i$ by rewriting equation \eqref{eq:xi}.
\begin{equation}\label{eq:quantum}
    X_i = 
	\Bigl|\sum_j e^{i(\theta_j-\phi_{ij})}\sqrt{X_jC_{j\mid i}}\Bigl|^2
\end{equation}
The phase factor of the observation $\theta_j$ is assumed to be observed together with the probability distribution $X_j$. Instead, the phase factors $\phi_{ij}$ are regarded as model parameters that can be learnt in the training process. The differences in the phase factors $\theta_j-\phi_{ij}$ control the interference between the interacting quantum states. 

\paragraph{Entropy.}
Consider for simplicity the classical probability in equation \eqref{eq:classic} where we split the summation in the set $S$ containing good predictors and the complementary set $S^c$ containing noise.
\begin{equation}
	X_{i} = \sum_{j \in S} C_{i\mid j} X_{j} + \sum_{j \in S^c} C_{i\mid j} X_{j}
\end{equation}
With a sufficiently large and balanced sample, the noise in $S^c$ would be equally distributed among the targets $i$, so that the summation over $S^c$ would not alter the ranking of the probabilities $X_{i}$. In practice, the finite sample size and the uncertainty affecting the entries of $C_{ij}$ may lead to a poor estimation of $X_{i}$ when the noisy features in $S^c$ are much more than the good predictors in $S$.
Entropy can be used to distinguish between signal/noise, and weight them accordingly. For each feature $j$ we compute the corresponding entropy with respect to the targets $i$.
\begin{equation}
	H_{j} = -\sum_{i}P_{i\mid j}ln(P_{i\mid j})
\end{equation}
The entropy $H_j$ is maximized for those features $j$ uniformly distributed across the targets $i$. However, these coincide with the less informative features only if the dataset is balanced. For this reason we introduce the joint probability $P_{ij}$ of the artificially balanced dataset that coincides with the conditional probability $C_{j\mid i}$ of the original dataset, up to a normalization constant.
Then, $P_{i\mid j}$ is the probability of $i$ given $j$ after balancing the dataset. 
\begin{equation}
P_{ij} = \frac{C_{ij}}{\sum_{j'} C_{ij'}}
,\quad
P_{i\mid j} = \frac{ P_{ij}}{\sum_{i'} P_{i'j}}
\end{equation}
In this setting, the configuration of zero entropy corresponds to perfect signal, while the configuration of maximum entropy corresponds to random noise. We introduce the following weights, giving weight 1 to perfect signal and weight 0 to random noise:
\begin{equation} \label{weight:entropy}
	\tilde H_{j} \equiv 1 - \frac{H_{j}}{H_{max}} = 1 + \frac{\sum_{i}P_{i\mid j}ln(P_{i\mid j})}{ln(\sum_{i} 1)}
\end{equation}
A robust estimator replaces the weights $C_{i\mid j}$ with $\tilde H_{j}C_{i\mid j}$. Introducing the robust weights in equation \eqref{eq:classic} and splitting again the summation in the sets $S$ and $S^c$, we get:
\begin{equation}
	X_{i} = \sum_{j \in S} \tilde H_{j}C_{i\mid j} X_{j} + \sum_{j \in S^c} \tilde H_{j}C_{i\mid j} X_{j}
	\approx \sum_{j \in S} \tilde H_{j}C_{i\mid j} X_{j}
\end{equation}
as $\tilde H_{j}$ is small for $j \in S^c$. In other words, the estimator is able to distinguish signal from noise and perform feature selection in a natural way. The same holds for the quantum probability in equation \eqref{eq:quantum} by replacing $C_{j\mid i}$ with $\tilde H_{j}^2C_{j\mid i}$.

\paragraph{Sparsity.} \label{sec:sparsity}
In real applications, both $C$ and $X$ in \eqref{eq:classic} or \eqref{eq:quantum} contain a large number of zero values. This is an issue both for the dimension of the tensors and because the product between $C$ and $X$ in \eqref{eq:classic} or \eqref{eq:quantum} is zero when all the non-zero entries of $X_{j}$ are associated with features $j$ that are never seen in the training set. In this case we obtain the degenerate probability estimate $X_{i}=0$. To solve this issue we define a fallback mechanism by applying a tensor contraction, i.e. we drop one dimension from the observation $X_{j_0...j_k...j_m} \rightarrow \sum_{j_k} X_{j_0...j_k...j_m}$ and compute the corresponding marginal distribution from the corpus $C_{i_0...i_nj_0...j_k...j_m} \rightarrow \sum_{j_k} C_{i_0...i_nj_0...j_k...j_m}$. 
We iterate until $X_{i}\neq 0$, eventually dropping all dimensions from $X_{j} \rightarrow X = 1$ and from the corpus $C_{ij} \rightarrow C_{i}$. In other words, in this particular case we estimate $X_{i}$ with the distribution through which the targets $i$ appeared in the corpus. 
We call \emph{policy} the order in which to drop the dimensions. In STC, the policy can be arbitrarily specified if prior knowledge is available, or learnt via reinforcement learning to identify the optimal order in which to drop the dimensions $j=(j_0,\dots,j_m)$ when the degenerate distribution $X_{i}=0$ is estimated. In this case, we want to identify the optimal sets $\{j_x\}_{x\in S_k}, ..., \{j_x\}_{x\in S_1}, \{\}$ to apply subsequently until estimating $X_{i}\neq 0$. 
We tackle the problem by implementing a myopic agent. In particular, we start from the state containing only the empty set $[\{\}]$ and we estimate $X_{i}\neq 0$ by construction for all the observations in the validation set. Then, given a loss function $L$,\footnote{Such as for instance the p-norm: $\frac{1}{2}\Bigl(\sum_{i} |X_{i}-\hat X_{i}|^p\Bigl)^{\frac{1}{p}}$}, we compute the reward $-L(X,\hat X)$ and the value of the state as the average reward received in all the iterations.
Subsequently, we add one index to the empty set and explore all the states $[\{\},\{j_0\}], ..., [\{\},\{j_m\}]$. In other words, we estimate $X_{i}$ by using only one of the indexes $j_0, j_1, \dots j_m$. We use $X_{i}$ obtained in the previous step for those observations resulting in $X_{i}=0$ and we compute the rewards for each state and the values of the states. Finally, we move to the state of maximum value $[\{\},\{j_*\}]$.
This procedure is repeated by adding one index to the current set of indices and exploring all the corresponding states until reaching the set $\{j_0,...,j_m\}$, containing all the indices, or when reaching a state where all the values of the next states are less than the value of the current state. 
The myopic agent learns the policy ${\{\}, \{j_x\}_{x\in S_1} ...,\{j_x\}_{x\in S_k}}$ that is sequentially improving the loss function $L$. We can now invert the order of the sets to understand which dimensions to drop first in case  $X_{i}=0$.
%The agent is implicitly computing feature importance and performing feature selection over the dimensions of the tensor. The scheme can be extended by allowing the agent to explore only the top states among those already explored in order to save computational time, by restricting the agent to complete a given subset of indices before trying the others in order to include prior knowledge in the learning, or by implementing a non-myopic agent for more flexibility. 

\paragraph{Tensorization.}
By representing the input data as a tensor, STC is natively suitable for working with multi-valued attributes and multi-labeled data (see e.g. \cite{yi2011multi}). In such cases, it may be convenient to represent the different attributes as separate dimensions of the tensor, in order to avoid mixing different aspects of the data items into a unique dimension of the feature space. When this is not required, for example when dealing with a single multi-valued attribute, such as text, or when dealing with multiple single-valued attributes, such as tabular data, STC can easily scale down to a matrix-based data representation, which uses a single dimension for all the features $j=(j_0)$. In this case the learning of the policy is not needed, as it reduces to $\{\}, \{j_0\}$ by construction.
%The capability of STC to deal with an arbitrary number of dimensions is not the main focus of this paper, as it may require additional efforts to learn the optimal shape of the tensor. Instead, we want to stress-test the quality and robustness of the standard configuration of STC that requires no data pre-processing or tuning. To this end, in Section~\ref{sec:test} we perform experiments that use a simple matrix-based data representation, at the additional advantage of avoiding the learning of the policy, which reduces to $\{\}, \{j_0\}$ by construction.

\section{Algorithm}\label{sec:algorithm}
In the training phase, we compute the corpus $C_{ij}$ given the set of observations in \eqref{eq:corpus}. Then, we set the phase factors $\phi_{ij}$ in equation \eqref{eq:predict}. These are regarded as model parameters that can be learnt from the data using gradient descent optimization algorithms \cite{ruder2016overview}, or arbitrary specified if prior knowledge is available. Also the policy can be arbitrary chosen or learnt from the data as described in Section \ref{sec:sparsity}.
In the prediction phase, given a new observation $X_{j}$ for which only the features $j$ are observed together with their phase factors $\theta_j$, we compute the normalized probability of the targets $\tilde X_i =X_i/\sum_i{X_i}$ with:
\begin{equation} \label{eq:predict}
	X_i = 
	\Bigl|\sum_j e^{i(\theta_j-\phi_{ij})}\tilde H^h_j \tilde C^p_{ij} X^p_j\Bigl|^\frac{1}{p}
	,\quad
	\tilde C_{ij} = \frac{C_{ij}}{(\sum_{i'} C_{i'j})^{1-b}(\sum_{j'} C_{ij'})^b}
\end{equation}
where $\tilde H_{j}$ is given in \eqref{weight:entropy} and where the entropy $h \geq 0$, the balance $b \geq 0$, and the power $p>0$ are the model hyper-parameters. If the estimated probability is degenerated (i.e., $X_{i}=0$), we apply repeatedly the policy chosen in the training phase.

\subsection{Properties}\label{sec:properties}
The {\bf classical probability} setting in \eqref{eq:classic} corresponds to the choice of the hyper-parameters $h=0$, $b=0$, $p=1$, when setting $\phi_{ij}=\theta_j=0$ in \eqref{eq:predict}. The {\bf quantum probability} setting in \eqref{eq:quantum} can be recovered with the particular choice $h=0$, $b=1$, and $p=\frac{1}{2}$. Choices different from the previous ones can be interpreted as a generalization of the probability measures where the power $p$ controls the probability amplitude. Smaller values of $p$ give similar weight to all the features regardless of their distribution. The entropic weights in \eqref{weight:entropy} are set with $h=1$, dropped with $h=0$, and their intensity can be tuned more in general with $h\geq 0$. The summation in \eqref{eq:predict} is dominated by fewer features for higher values of the entropy $h$, that lead to predictions based on less but more relevant features, thus more {\bf robust to noise}. Missing values can be either ignored or treated like any other categorical value with their own semantics. The second approach allows to uncover patterns among the {\bf missing data}. STC deals with {\bf imbalanced data} by artificially balancing the sample when setting $b=1$ in equation \eqref{eq:predict}. Here $b=0$ uses the conditional probability of $i$ given $j$ that is affected by imbalanced data, while $b=1$ uses the conditional probability of $j$ given $i$ that artificially balances the sample. For $0<b<1$ the sample is semi-balanced, increasing the weight of the less frequent classes but not enough to have a balanced sample. For $b>1$ the sample is super-balanced, where the weight of the less frequent classes is greater than the weight of the most frequent classes.
The STC algorithm can be turned into a native {\bf multilabel classifier} by selecting the indices with highest probability $i^*_0...i^*_n = \arg\max X_{i_0...i_n}$, thus predicting several targets at the same time.
The algorithm is ready to use in an {\bf online learning} context. As new data $X_{ij}$ become available, they can be added to the corpus with minimal computational effort, that is $C_{ij} + X_{ij}$. STC supports a purely data-driven approach by estimating the optimal policy via reinforcement learning, learning the parameters $\phi_{ij}$ with optimization algorithms, and tuning the hyper-parameters via cross-validation. However, it is also possible to include {\bf prior knowledge} in the model by arbitrarily specifying the policy, arbitrarily setting the phase factors $\phi_{ij}$, and arbitrarily choosing the hyper-parameters $h$, $b$, $p$ to recover the classical or quantum probability. A mixed approach is also possible. 

\subsection{Explainability} \label{sec:explanation}

\paragraph{Local explanation.} For what concerns the explanation of predictions on single instances (local explanation), the contribution of each feature $j$ to the total probability $X_i$ is given by the addend $e^{i(\theta_j-\phi_{ij})}\tilde H^h_j \tilde C^p_{ij} X^p_j$ in \eqref{eq:predict}. Its modulus can be used to measure the importance of the features and assess the degree in which they contribute to the classification. In other words, the most
influential feature for the classification $i^*$ is given by:
\begin{equation}
	j^* = \arg\max_{j}\tilde H^h_j \tilde C^p_{i^*j} X^p_j
\end{equation}
$\tilde H^h_j \tilde C^p_{i^*j} X^p_j$ is used in general to rank the features $j$ by importance. Moreover, the phase factor $e^{i(\theta_j-\phi_{i^*j})}$ gives the direction of the contributions in the unit circle and it is able to uncover constructive or destructive interference between the features and the classification $i^*$.

\paragraph{Global explanation.} The explanation at the class level (global explanation) is obtained by investigating the tensor $\tilde H^h_j \tilde C^p_{ij}$ to detect the global most influential feature for each target $i$ according to:
\begin{equation}
	j_i^* = \arg\max_{j}\tilde H^h_j \tilde C^p_{ij}
\end{equation}
$\tilde H^h_j \tilde C^p_{ij}$ is used in general to rank the features $j$ by their global importance with respect to target $i$. Moreover, when using $h\neq 0$, the features with best discriminatory power among all the targets are given by the largest entries of $\tilde H^h_j$. Finally, the phase factor $e^{i\phi_{ij}}$ is used to uncover the interference pattern underlying the system. 

\section{Empirical evaluation}\label{sec:test}
The goal of our empirical evaluation of STC is to assess the classification performance and the interpretability of the explanation, without relying on data pre-processing, (hyper-)parameter tuning, or learning of the policy. 
The capability of STC to deal with an arbitrary number of dimensions is not the main focus of this paper, as it may require additional efforts to learn the optimal shape of the tensor and the corresponding policy. 
Instead, we want to stress-test the quality and robustness of the standard configuration of STC. To this end, we perform experiments that use a simple matrix-based data representation as discussed in Section \ref{sec:theory} and
we execute STC with entropic weights by relying both on classical probability ({\small\sf STC Classic (STC-C)}) and on quantum probability ({\small\sf STC Quantum (STC-Q)}) as described in Section \ref{sec:properties}.\footnote{{\small\sf STC-C}: $h=1$, $b=0$, $p=1$. {\small\sf STC-Q}: $h=1$, $b=1$, $p=1/2$.} Finally, we limit STC to real-valued wave functions by ignoring all the phase factors in \eqref{eq:predict} $\phi_{ij}=\theta_j=0$. 

In all the experiments, our approach is to compare STC against six consolidated algorithms, namely {\small\sf Decision Tree (DT)}, {\small\sf Random Forest (RF)}, {\small\sf K-Nearest Neighbors (KNN)}, {\small\sf Support Vector Machine (SVM)}, {\small\sf Multinomial Naive Bayes (MNB)}, and {\small\sf Logistic Regression (LR)}. The choice is motivated by the purpose of covering a spectrum of very diverse solutions for supervised classification and to exploit a homogeneous and stable third-part implementation using {\small\sf scikit-learn} \cite{scikit-learn}. In the second experiment, we also include evaluation performances of recent deep learning approaches as reported in literature.

All the experiments have been executed on a HP Z420 Workstation with Intel(R) Xeon(R) CPU 3.60GHz and 16GB System Memory equipped with Ubuntu 14.04. We also run all the experiment code on MacOS Big Sur and Windows 10. All the code is written in Python 3.7.

\subsection{Structured data}
In our first experiment, we rely on the {\small\sf Zoo} dataset~\cite{Dua:2019}.
%\footnote{See https://archive.ics.uci.edu/ml/datasets/Zoo} 
{\small\sf Zoo} provides a structured representation of 101 animals featured by 16 attributes. All the attributes, except for the number of legs, describe animal characteristics as boolean values denoting the fact that the animal has the characteristic at hand. For the number of legs we have 6 possible values. In the target there are 7 animal classes (e.g., {\small\sf Mammal}, {\small\sf Bird}, etc.). The dataset is quite imbalanced with the larger class ({\small\sf Mammal}) containing 41 animals and the smaller ({\small\sf Amphibian}) only 4 animals. The choice of {\small\sf Zoo} is motivated by the fact that it is an interesting case where the number of instances is relatively small with respect to the number of classes and where classes are imbalanced.
In the experiment, we execute STC and its competing algorithms in 100 independent runs. For each run the composition of the training and the test sets is randomly selected, with the test set size equal to the 30\% of the whole dataset. For the competing algorithms, in each run we set the hyper-parameters through model selection performed by the grid search strategy on the training set targeting the weighted average f1-score. The model selection procedure includes the option of scaling data with respect to the minimum and maximum values. Instead, for STC we keep the standard configuration in order to stress-test our methodology.

The average precision, recall, and f1-score weighted by the class support over the 100 runs are reported in Table~\ref{tab:zooavg}, together with macro-averages. 
In this experiment, we are also interested in studying the stability of the algorithms with respect to the variation of the training set. To this end, we execute a pairwise comparison of the algorithms by counting, for each pair of algorithms, in how many experiment runs one of the two over-performed the other in terms of weighted f1-score. These results are also reported in Table~\ref{tab:zooavg}, where each entry $[i,j]$ of the table contains the fraction of runs in which the $i$-th algorithm achieved a better f1-score than the $j$-th competitor (excluding ties).

\begin{table}[!htp]
\caption{Average precision, recall, f1-score weighted by the class support, f1-macro, and pairwise competition over 100 runs of the {\small\sf Zoo} experiment.}
\begin{center}
\begin{tabular}{cc}
{\tiny\sf
\begin{tabular}{rcccc}
\toprule
{} &  Prec. &  Rec. &  F1 &  F1-macro \\
\midrule
DT & 0.944 & 0.935 & 0.931 &  0.836\\
RF & 0.936 & 0.931 & 0.923 &  0.827\\
KNN & 0.916 & 0.915 & 0.905 &  0.792\\
SVM & 0.946 & 0.941 & 0.934 &  0.847\\
MNB & 0.916 & 0.903 & 0.897 &  0.764\\
LR & 0.946 & 0.938 & 0.932 &  0.843\\
\midrule
STC-C & 0.790 & 0.850 & 0.804 &  0.659\\
STC-Q & \textbf{0.953} & \textbf{0.953} & \textbf{0.945} & \textbf{0.871} \\
\bottomrule
\end{tabular}}
\;
{\tiny\sf
\begin{tabular}{cccccc|cc|c}
\toprule
DT &    RF &   KNN &   SVM &   MNB &    LR &  STC-C &  STC-Q &  Mean \\
\midrule
0.00 &  0.61 &  0.73 &  0.50 &  0.80 &  0.54 &   0.97 &   0.38 &  0.57 \\
0.39 &  0.00 &  0.70 &  0.32 &  0.72 &  0.38 &   0.98 &   0.34 &  0.48 \\
0.27 &  0.30 &  0.00 &  0.29 &  0.63 &  0.28 &   0.96 &   0.25 &  0.37 \\
0.50 &  \textbf{0.68} &  0.71 &  0.00 &  0.79 &  0.52 &   0.96 &   \textbf{0.39} &  0.57 \\
0.20 &  0.28 &  0.37 &  0.21 &  0.00 &  0.22 &   0.86 &   0.13 &  0.28 \\
0.46 &  0.62 &  0.72 &  0.48 &  0.78 &  0.00 &   0.96 &   \textbf{0.39} &  0.55 \\
\midrule
0.03 &  0.02 &  0.04 &  0.04 &  0.14 &  0.04 &   0.00 &   0.00 &  0.04 \\
\textbf{0.62} &  0.66 &  \textbf{0.75} &  \textbf{0.61} &  \textbf{0.87} &  \textbf{0.61} &   \textbf{1.00} &   0.00 &  \textbf{0.64} \\
\bottomrule
\end{tabular}}
\\
\end{tabular}
\end{center}
\label{tab:zooavg}
\end{table}

\paragraph{Results.}
We note how the quantum approach of {\small\sf STC-Q} outperforms all the competing algorithms in Table \ref{tab:zooavg} but also demonstrates to be the most stable with respect to the variation of the training set. {\small\sf STC-Q} is always the best choice against all the competitors and outperforms all the other algorithms in more than 50\% of the runs.
The comparison between {\small\sf STC-Q} and {\small\sf STC-C} highlights the benefits of switching from the classical to the quantum probability. 
%In general, we observe that {\small\sf STC-Q} performs well in this dataset where the number of instances is relatively small with respect to the number of classes and where classes are imbalanced.

\subsection{Text classification}
In our second experiment, we want to evaluate the capability of STC to deal with unstructured and noisy data. To this end, we execute a task of text classification on the well-known {\small\sf 20 Newsgroups} dataset.\footnote{We use the \emph{by-date} version of the dataset obtained from
\url{http://qwone.com/~jason/20Newsgroups/}} {\small\sf 20 Newsgroups} provides about 19 000 news divided in a training and a test set and organized in 20 thematic classes that are almost balanced in terms of number of instances. In order to keep the text as noisy as possible and stress-test the algorithms, we perform only a simple word tokenization by the function {\small\sf nltk.word\_tokenize}.\footnote{See https://www.nltk.org/book/} No other text transformation and cleaning procedure is performed in order to keep all the morphological and syntactical variety of text, including plurals, capital letters, stop-words and punctuation. The final dataset is composed by 185 733 features (i.e., document terms) and 11 314 documents in the training set and 7 532 documents in the test set.
One of the main ideas of this experiment is to test what happens when a user takes a supervised classification algorithm off the shelf and applies it to a text dataset without spending any effort in tuning, feature engineering, or data pre-processing. To simulate this situation, we do not perform model selection and we use the default hyper-parameter values for STC competitors as they are defined in {\small\sf scikit-learn}. Also for text vectorization and term weighting we rely on Tf-Idf as one of the most popular and standard approaches for representing text data in text classification. For what concerns STC, we exploit the standard configuration of {\small\sf STC-Q}, which works directly on terms without needing any vectorization or weighting scheme. In order to include an example of generalized probability without performing any kind of model selection, we also explore the results obtained by modifying the quantum probability amplitude of {\small\sf STC-Q} from the square to the cubic root, thus setting the hyper-parameter $p=1/3$ instead of $p=1/2$. We extend the comparison to the performances reported in literature of recent deep learning approaches to text classification on {\small\sf 20 Newsgroups}, including {\small\sf TextEnt} \cite{yamada2018}, the Neural Attentive Bag-of-Entities Model ({\small\sf NABoE}) \cite{yamada2019}, Cooperative Neural Networks ({\small\sf CoNN}) \cite{shrivastava2018}, and the Diversified Ensemble Neural Network ({\small\sf DEns}) \cite{zhang2020}. The results are shown in Table~\ref{tab:news}, where we report the macro average of the evaluation metrics in order to be compliant with the results reported in literature.

\begin{table}[!ht]
\caption{Comparison between STC and other text classification algorithms on average macro precision (Prec.), recall (Rec.), f1-score (F1), and accuracy (Acc.)}
\begin{center}
{\tiny\sf
    \begin{tabular}{r|cccccc|c}
                &  DT & RF & KNN & SVM & MNB & LR & STC-Q \\
                \toprule
      Prec.     &  0.546 & 0.748 & 0.599 & 0.799 & 0.822 & 0.808 & \textbf{0.863} \\
      Rec.     &  0.543 & 0.727 & 0.528 & 0.779 & 0.725 & 0.795 & \textbf{0.856} \\
      F1     &  0.543 & 0.725 & 0.539 & 0.784 & 0.724 & 0.797 & \textbf{0.856} \\
      Acc.     &  0.549 & 0.739 & 0.529 & 0.788 & 0.744 & 0.805 & \textbf{0.864} \\
      \bottomrule
      \multicolumn{5}{r}{}\\
    \end{tabular}
    \quad
    \begin{tabular}{cccc|c}
                NABoE & TextEnt & CoNN & DEns & STC-Q$^{*}$ \\
                \toprule
      N/A & N/A & N/A & N/A & \textbf{0.871} \\
      N/A & N/A & N/A & N/A & \textbf{0.866} \\
      0.862 & 0.839 & N/A & N/A & \textbf{0.866} \\
      0.868 & 0.845 & 0.837 & 0.871 & \textbf{0.873} \\
      \bottomrule
      \multicolumn{5}{r}{$\scriptscriptstyle ^{*}$Executed by setting p = 1/3}\\
    \end{tabular}
}
\end{center}
\label{tab:news}
\end{table}%

The computational time required by STC is comparable with {\small\sf Random Forest} for training (about 50s) and with {\small\sf Support Vector Machine} for prediction (about 80s). 

\paragraph{Results.} 
The results reported in Table~\ref{tab:news} show that the standard configuration of STC-Q significantly outperforms the default configurations of the competing algorithms. 
Remarkably, STC-Q achieves results comparable, if not superior, to the deep learning approaches and improves the state-of-the-art. The result is especially surprising when considering that the deep learning approaches are specifically tuned for the given task, involve non-trivial data pre-processing, exploit semantics, embeddings and external knowledge bases, while our simple methodology avoids any text transformation and tuning.
%The results reported in Table~\ref{tab:news} show that the standard configuration of STC-Q outperforms the competing algorithms and achieves results comparable with the deep learning approaches. Remarkably, the performances with $p=1/3$ improve the state-of-the-art. The result is especially surprising when considering that the deep learning approaches are specifically tuned for the given task, involve non-trivial data pre-processing, exploit semantics, embeddings and external knowledge bases, while our simple methodology avoids any text transformation and tuning.

%The most relevant remark about the experiment is that this performance is obtained with the default configuration of STC and, most relevant, without any text transformation. In text classification, achieving good results working on the text as-is is highly desirable property in that it can significantly reduce the work required in real applications. With respect to the results reported in literature, we stress the fact that we compare STC against algorithms that have been specifically tuned for text classification and, in many cases, exploit text transformation to achieve their results.

\subsection{Explanation}
In our third experiment, we want STC to explain the state-of-the-art results obtained on {\small\sf 20 Newsgroups} ($p=1/3$). In particular, we want to evaluate how much the terms in the native explanation of STC are interpretable, that is how much they appear to be meaningful for a human evaluator.
We focus the experiment on global explanation by involving a small group of 45 anonymous evaluators in a survey. Each evaluator receives the same set of document terms extracted from {\small\sf 20 Newsgroups} and is asked to decide whether each term is {\small\sf Relevant}, {\small\sf Neutral} or {\small\sf Not Relevant} for a given class in {\small\sf 20 Newsgroups}.
%\footnote{In order to keep the survey simple, we selected only the classes rec.sport.hockey, rec.sport.baseball, misc.forsale, sci.crypt, soc.religion.christian, and talk.politics.mideast} 
The terms submitted to evaluation have been chosen according to the following procedure. First, for what concerns the competing algorithms, we exploit LIME \cite{ribeiro2016should} in that it provides a uniform and model agnostic method for producing local relevance scores of terms for each instance in the test set with respect to the predicted class. Then, we create the global scores of terms for each algorithm by taking the sum of their local relevance scores for each class.
To allow for a natural comparison, for STC we compute the relevance scores by summing the local weights described in Section \ref{sec:explanation} of all the instances in the test set.
For each class in the survey, the final list has been composed by merging the top 6 relevant terms in each algorithm explanation (hiding the provenance) into a unique list. From this list, we removed English stopwords, punctuation, unclear abbreviations and other terms that are clearly unrelated to the target class in order to submit a list of about 25 terms per class to the human evaluators.
The result is evaluated as follows. Given a class $c$ and a term $t$, we denote $r_{c}(t)$ the fraction of the evaluators who evaluated $t$ being relevant for $c$ and $n_{c}(t)$ the fraction who evaluated $t$ as not relevant. The score of $t$ for $c$ is then $\sigma_c(t) = r_c(t) - n_c(t)$ with $\sigma_c(t)=-1$ for terms removed from the survey list. The final evaluation score for the algorithm $i$ is the mean of the human evaluations over all the terms produced by $i$ for all the classes in the survey.

\textbf{Results.}
We find that STC achieves a score of $0.598$, corresponding to a better interpretability than {\small\sf RF} ($0.376$), {\small\sf SVM} ($0.270$), {\small\sf LR} ($0.256$), {\small\sf DT} ($0.146$), {\small\sf KNN} ($-0.200$). In this sense, STC is similar to one of the most well-known and interpretable models: {\small\sf MNB} ($0.696$). We also notice that the results achieved by all the competing algorithms are due to the exploitation of LIME, which is remarkably inefficient in terms of computation time (about 30s for each document, resulting in more than 60 hours for each algorithm). Next, we extend the comparison with {\small\sf MNB} by exploring the probability $P(t \mid c)$ of a term $t$ given the class $c$ as internally computed by the model. We compare this explanation provided by {\small\sf MNB} natively ({\small\sf MNB-G}) with the STC global explanation ({\small\sf STC-G}) computed as described in Section \ref{sec:explanation}. A short example of the 6 top terms for the class {\small\sf rec.sport.baseball} is shown in Table~\ref{tab:lime}, where we also report the top terms obtained by aggregating the native local STC weights and those obtained with {\small\sf MNB} through LIME.
It is interesting to note that the two global explanations derived from local explanations (i.e., {\small\sf STC} and {\small\sf MNB}) are similar, while {\small\sf STC-G} provides very specific and class-related terms by selecting baseball teams and players names, such as in an entity recognition system. We note this behavior also for the other classes. These specific terms are less interpretable, in that the survey shows that many of them are not known to evaluators, but potentially more useful to explore a class content. The native explanation of {\small\sf MNB-G} is instead dominated by noise, that STC has been able to filter out by exploiting the entropic weights described in Section \ref{sec:theory}.

\begin{table}[!ht]
\caption{Top-6 terms in the explanations of STC and MNB for the class {\small\sf rec.sport.baseball} obtained by aggregating the local weights (STC \& MNB) or computed natively (STC-G \& MNB-G).}
\begin{center}
\begin{tabular}{cc}
    {\tiny\sf 
    \begin{tabular}{c|l}
    \toprule
        STC & ``game", ``team", ``baseball", ``games", ``he", ``pitcher" \\
        MNB & ``pitcher", ``baseball, "game, ``Sox", ``games", ``Braves" \\
\bottomrule
    \end{tabular}
    }
&
    {\tiny\sf 
    \begin{tabular}{c|l}
    \toprule
        STC-G & ``Phillies", ``pitching", ``Braves", ``Alomar", ``Mets", ``Players" \\
        MNB-G & ``$>$", ``:", ``," , ``the", ``.", ``--" \\
        \bottomrule
    \end{tabular}
    }
\\
\end{tabular}
\end{center}
%begin{center}
%{\small\sf
%\begin{tabular}{cc|cc}
%\toprule
%STC	&	MNB	& STC-G	&	MNB-G \\
%\toprule
%game	&	pitcher	&	Phillies	&	$>$ \\
%team	&	baseball	&	pitching	&	:	\\
%baseball	&	game	&	Braves	&	,\\
%games	&	Sox	&	Alomar	&	the\\
%he	&	games	&	Mets	&	.\\
%pitcher	&	Braves	&	Players	&	--\\
%\bottomrule
%\end{tabular}}
%\end{center}
\label{tab:lime}
\end{table}%

\section{Concluding remarks}\label{sec:conc}
By regarding an observation as a superposition of features, we introduce the concept of wave-particle duality in machine learning and present Sparse Tensor Classifier (STC). 
The empirical evaluation against consolidated algorithms and deep learning alternatives demonstrates the native capability of STC to achieve state-of-the-art performances without relying on data pre-processing and hyper-parameter tuning, at the additional benefit of providing a meaningful explanation of the classification results.
Our future work will be devoted to further investigating STC from complementary perspectives. In particular, we aim at developing a learning procedure for the phase factors, investigating their implications in terms of explainability and classification performance, assessing the benefits of a tensor representation of the data items, and exploring the extension of STC to work with ordinal and continuous variables.
We hope our novel methodology illustrates how the analogy with the theory of superposition of states in quantum physics can be successfully exploited in machine learning and open the door to future research in this direction.

\bibliography{stc}
\bibliographystyle{plain}

\end{document}